\pdfoutput=1

\documentclass[11pt]{article}

\usepackage{ACL2023}

\usepackage{times}
\usepackage{latexsym}

\usepackage{amsmath}
\usepackage{array}

\usepackage{graphicx}
\usepackage{algorithm}
\usepackage{algorithmic}
\usepackage{booktabs}
\usepackage{multirow}
\usepackage{amsfonts}
\usepackage{amssymb}

\usepackage{nicefrac}
\usepackage{url}

\makeatletter
\newcommand{\ssymbol}[1]{$^{\@fnsymbol{#1}}$}
\makeatother

\newcommand{\myparagraph}[1]{\textbf{#1}\hspace{1.8ex}}

\usepackage[T1]{fontenc}
 
\usepackage[utf8]{inputenc}

\usepackage{microtype}

\usepackage{inconsolata}

\title{Multimodal Prompt Learning for Product Title Generation \\ with Extremely Limited Labels}

\author{Bang Yang\textsuperscript{1}\thanks{\ \ Equal Contributions. Ordered by a coin toss.}, Fenglin Liu\textsuperscript{2}\footnotemark[1], Zheng Li\textsuperscript{3}\thanks{\ \ Corresponding authors.}, Qingyu Yin\textsuperscript{3}, \\ \bf Chenyu You\textsuperscript{4}, Bing Yin\textsuperscript{3}, Yuexian Zou\textsuperscript{1}\footnotemark[2] 
\\
\textsuperscript{1}School of ECE, Peking University, China \ \
\textsuperscript{2}University of Oxford, United Kingdom \\ 
\textsuperscript{3}Amazon.com Inc, Palo Alto, USA \ \
\textsuperscript{4} Yale University, USA\\
{\tt \{yangbang, zouyx\}@pku.edu.cn;  fenglin.liu@eng.ox.ac.uk}  \\ 
 {\tt chenyu.you@yale.edu; \{amzzhe, qingyy, alexbyin\}@amazon.com}\\ 
}

\begin{document}
\maketitle
 
\begin{abstract}
   Generating an informative and attractive title for the product is a crucial task for e-commerce. Most existing works follow the standard multimodal natural language generation approaches, e.g., image captioning, and employ the large scale of human-labelled datasets to train desirable models. However, for novel products, especially in a different domain, there are few existing labelled data. In this paper, we propose a prompt-based approach, i.e., the Multimodal Prompt Learning framework, to accurately and efficiently generate titles for novel products with limited labels. We observe that the core challenges of novel product title generation are the understanding of novel product characteristics and the generation of  titles in a novel writing style. To this end, we build a set of multimodal prompts from different modalities to preserve the corresponding characteristics and writing styles of novel products. As a result, with extremely limited labels for training, the proposed method can retrieve the multimodal prompts to generate desirable titles for novel products. The experiments and analyses are conducted on five novel product categories under both the in-domain and out-of-domain experimental settings. The results show that, with only 1\% of downstream labelled data for training, our proposed approach achieves the best few-shot results and even achieves competitive results with  fully-supervised methods trained on 100\% of training data; With the full labelled data for training, our method achieves state-of-the-art results.
\end{abstract}

\begin{figure}[t]
\centering
\includegraphics[width=0.98\linewidth]{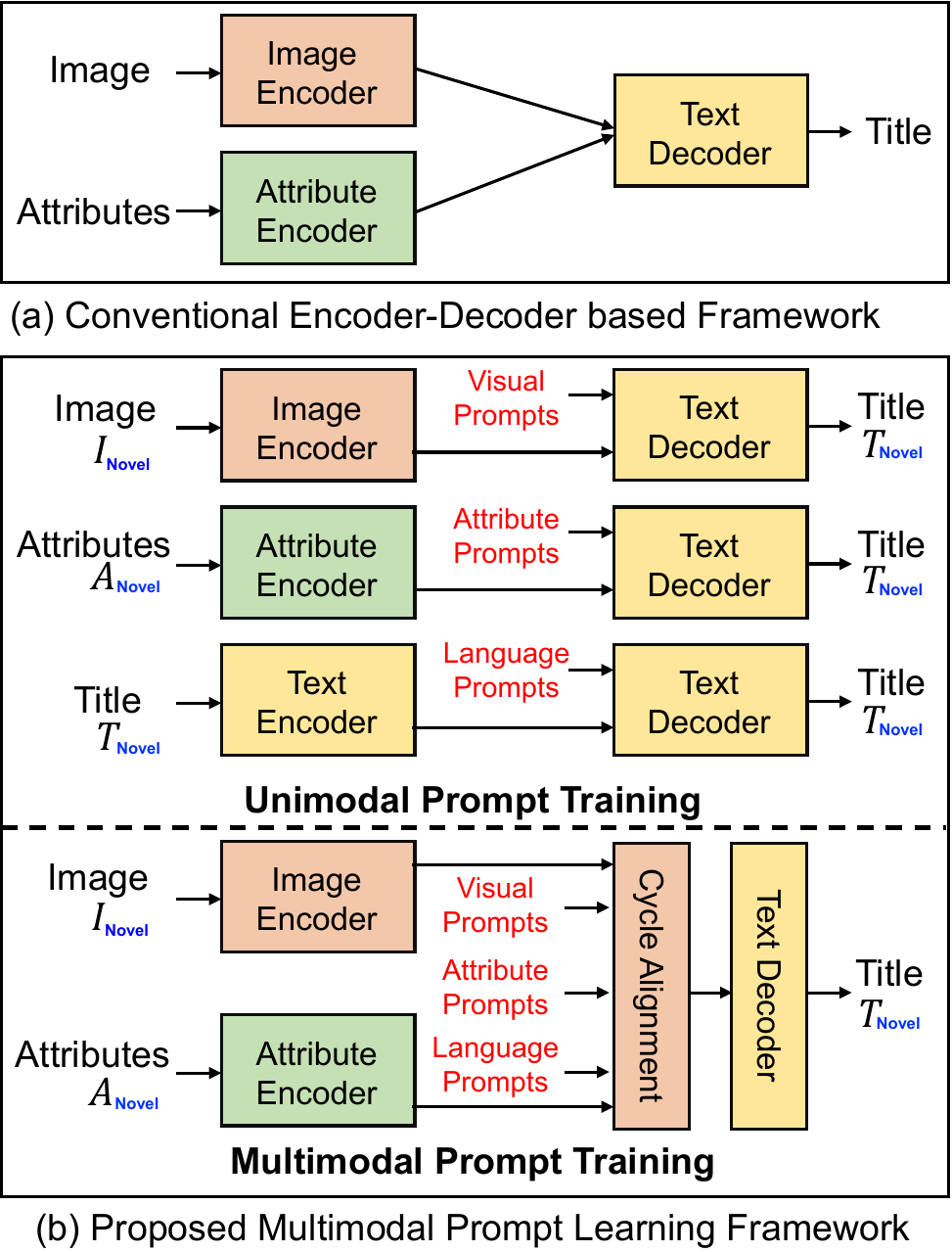}
\caption{(a) The conventional encoder-decoder-based generation framework used for product title generation. (b) Our proposed Multimodal Prompt Learning framework first introduces the unimodal prompt training to learn the domain characteristics and writing styles of novel products, which can be encoded in the trainable prompts across different modalities, and then introduces the multimodal prompt training to highlight and capture the important characteristics from prompts for generating accurate novel product titles with limited data.}
\label{fig:introduction}
\end{figure}

\section{Introduction}
\label{sec:intro}

Product title generation aims to comprehend the content of a given product provided by merchants, which may come in various forms such as an input product image and a set of attributes, and then automatically generate an appealing and informative title.
The generated title should contain essential product characteristics, along with the product details, e.g., brand name, category, style, size, material, and colour \cite{Song2022Product,mane2020product,zhan2021probing}.
Therefore, a desirable title can highlight the characteristics and advantages of the product, leading to time savings for consumers, enhancing their overall shopping experience, and ultimately increasing product sales.
Admittedly, in E-commerce, the ability to perform product title generation automatically offers the possibility of relieving merchants from the time-consuming analysis of complex product details and writing concise and appealing titles; and alerting merchants of important product characteristics and advantages \cite{chen2019towards,Zhang2019Product,de2018generating,zhang2019automatic}.

In general, the task of product title generation can be defined as a data-to-text problem.
Following existing efforts on data-to-text tasks \cite{specia2016shared,hossain2019comprehensive,Bahdanau2015NMT}, Figure~\ref{fig:introduction}(a) shows the conventional product title generation approach: the encoder-decoder framework.
The image encoder and attribute encoder respectively transform the product image and product attributes into visual and attribute representations, which the text decoder subsequently decodes into a product title. 
Such encoder-decoder-based methods have achieved great success in advancing the state-of-the-art of various data-to-text tasks, e.g., image captioning \cite{hossain2019comprehensive,Shan2022ERNIE}, multimodal machine translation \cite{specia2016shared}, and video captioning \cite{Yang2021NACF,Yu2016Hierarchical}.
However, these methods rely on a large volume of annotated data, which is particularly time-consuming to collect. 
This issue is especially severe in the E-commerce title generation scenario, where products from different categories always contain category-specific attributes. Therefore, the product title generation model trained on existing products cannot be directly used on novel products, such as with new categories or new designs. Nevertheless, it is difficult to collect and label sufficient training data in a timely manner, which prevents the rapid deployment of such encoder-decoder models online.

As shown in Figure~\ref{fig:introduction}(b), we propose the Multimodal Prompt Learning (MPL) framework, which deals with the situation where the training data is scarce. 
In detail, we observe that novel product title involves different domain product characteristics (e.g., category-specific attributes) and different writing styles, directly adopting a model or transferring a model pre-trained on existing available product data to novel product data will significantly degrade the performance, especially when the labelled data (i.e., image-attribute-title pairs) is insufficient in quantity \cite{Wang2019Characterizing}.
To this end, we first construct a set of multimodal prompts from different modalities, i.e., visual prompts, attribute prompts, and language prompts.
During training, given the limited data of novel products (i.e., Image $I$ - Attribute $A$ - Title  $T$), to make full use of it, MPL introduces the unimodal prompt training  to enable the different prompts to preserve the corresponding domain characteristics and the writing styles of novel products from different modalities/perspectives.
In implementations, 
(i) we introduce the visual prompts $\mathcal{P}_I$ to train the model by generating the title $T$ in the $I \to \mathcal{P}_I \to T$ pipeline; 
(ii) we introduce the attributes prompts $\mathcal{P}_A$ to train the model in the $A \to \mathcal{P}_A \to T$  pipeline;
(iii) we introduces the textual language prompts $\mathcal{P}_T$ to train the model by reconstructing the title $T$ in the $T \to \mathcal{P}_T \to T$ auto-encoding pipeline.
It is worth noting that the auto-encoding pipeline aims to reconstruct the same input sentence, therefore, it is straightforward for the model to be trained \cite{Wang2016Auto,Tschannen2018Recent} to learn the necessary domain characteristics and the writing styles of novel products via the small amount of data. 
Besides, the unsupervised auto-encoding process provides opportunities for our model to be further improved by incorporating more unlabelled text-only data \cite{nukrai2022text}.
At last, MPL introduces multimodal prompt training to learn to generate accurate novel product titles with the help of learned multimodal prompts.
In the implementation, we first introduce a Cycle Alignment Network to highlight and capture the important characteristics from multiple modalities by cycle aligning three types of prompts; then take the input images $I$ and attributes $A$ of novel products as queries to retrieve the learned domain characteristics in the aligned prompts; and finally rely on the learned writing styles in the text decoder to generate the titles for the novel products.

In this way, the proposed MPL framework can accurately and efficiently generate novel product titles with limited training data by 1) introducing multimodal prompts to learn domain characteristics and writing styles of novel products;
2) learning to accurately highlight the product characteristics and advantages across multiple modalities.
It enables our approach to be rapidly well-adapted to the novel product domain, helping sellers save time in deploying new products, optimizing consumers' consumption experience, and thus boosting sales. 
The experiments and analyses on a large-scale dataset, i.e., Amazon Product Dataset \cite{Ni2019Amazon}, across five novel product categories prove the effectiveness of our approach.

Overall, the contributions are as follows:
\begin{itemize}
    \item We propose the  Multimodal Prompt Learning (MPL) framework to generate few-shot novel product titles, where the training data in the novel product domain is scarce.
    
    \item Our MPL framework first introduces multiple types of prompts to learn the domain characteristics and writing styles of novel products, and then learns to generate accurate final titles by highlighting and capturing the important characteristics from multiple modalities.
    
    \item Our experiments on five novel products prove the effectiveness of our approach, which generates desirable product titles for novel products with only 1\% of the training data otherwise required by previous methods, and significantly outperforms state-of-the-art results with the full training data.
\end{itemize}

\section{Related Work}
The related works are discussed from 1) Product Description and 2) Few-shot Learning.

\subsection{Product Description}
Generating the product titles to describe the given products is similar to the multimodal language generation tasks, e.g., image captioning \cite{Xu2015Show,chen2015microsoft,fenglin2019MIA} and multimodal machine translation \cite{specia2016shared}.
To perform multimodal language generation tasks, a large number of encoder-decoder-based models have been proposed \cite{Guo2022LVP,Zhang2023Universal,Shan2022ERNIE,Yang2021NACF,chen2015microsoft,anderson2018bottom,yang2018SGAE,Cornia2020M2,fenglin2020aimNet,10095809,zhu2023},
in which a CNN \cite{Krizhevsky2012CNN} and a LSTM/Transformer \cite{Hochreiter1997LSTM,Vaswani2017Transformer,fenglin2020rskip} is used as the image encoder and text encoder to encode the input images and texts, and an LSTM \cite{Hochreiter1997LSTM} or a Transformer \cite{Vaswani2017Transformer,fenglin2020rskip} is used as the text encoder to generate the final sentences.
Inspired by the great success of an encoder-decoder framework in multimodal language generation tasks, existing efforts on product description have proposed a wide variety of encoder-decoder based frameworks \cite{Song2022Product,Zhang2019Product,mane2020product,zhan2021probing,chan2020selection,zhang2019automatic,gong2019automatic,de2018generating,chen2019towards} to describe given products.
However, these existing models are trained on large-scale datasets, while collecting data on novel products, e.g., novel categories and novel designs, to train the models is typically very limited.
To this end, we propose multimodal prompt learning to relax the reliance on the training dataset for the few-shot novel product description - with the goal of quick deployment of new products.

\subsection{Few-shot Learning}
Recently, few-shot learning \cite{Wang2020Fewshot} has received growing research interest across many AI domains \cite{Dhillon2020fewshot,Tian2020fewshot, Perez2021True,Gu2022PPT,Gao2021Making,Tsimpoukelli2021Multimodal,Zha2022Disentangling,Wang2022Learning,Li2021MetaTS,Huang2022Multilingual,Wang2022RETE,Li2020Learn,Zhang2021QUEACO}. Inspired by the success of few-shot learning, several works \cite{liu2021relative,sreepada2020mitigating,gong2020learning,Zhou2022Few-Shot,xu2021fewclue} explored such an approach for the domain of E-commerce.
However, most focus on unimodal tasks, either on the graph data (e.g., node classification, recommendation) \cite{liu2021relative,sreepada2020mitigating,Wang2022Learning,Li2020Learn,Wang2022RETE,Huang2022Multilingual}, or on the text data (e.g., sentiment analysis and recommendation) \cite{gong2020learning,xu2021fewclue,Zha2022Disentangling}, or on the image data (e.g., image classification) \cite{Zhou2022Few-Shot}.
As a multimodal task incorporating disparities between the visual and the textual modalities \cite{Liang2022Mind}, few-shot product title generation is far more challenging.
To prove our hypothesis, we re-implement existing few-shot learning methods for novel product title generation, demonstrating with our experiments that our approach significantly outperforms existing methods.

\section{Approach}
In this section, we will introduce the proposed Multimodal Prompt Learning  (MPL) method in detail.

\subsection{Formulation}
Given the basic product information, i.e., product image $I$ and product attribute $A$, the goal of product title generation is to generate an accurate and concise product title $T=\{w_1,w_2,\dots,w_{N}\}$, including $N$ words. 
Current state-of-the-art methods usually consist of an image encoder and a text encoder to extract the image representations $R_I$ and attribute representations $R_A$, and a text decoder to generate the target title $T$, which is formulated as:
\begin{equation}
\label{eq:conventional}
\left\{
\begin{array}{ll}
\text{Image Encoder}  : I \to R_I ; \\
\text{Attribute Encoder}  : A \to R_A ; \\
\text{Text Decoder} : \{R_I, R_A\} \to T .
\end{array}
\right.
\end{equation}
Existing works rely on the annotated data image-attribute-title pairs to train the model by minimizing a supervised training loss, e.g., cross-entropy loss.
However, for many novel products, only a small amount of data is available.
In this case, we have to collect sufficient data to train the model, while collecting and labelling data is particularly labour-intensive and expensive.
As a result, insufficient training data poses a great challenge for building models to describe novel products.

To this end, we propose the MPL generation framework to generate accurate and desirable titles when encountering a novel product.
MPL includes two components: Unimodal Prompt Training (UPT) and Multimodal Prompt Training (MPT), where the former introduces three types of prompts (visual prompts $\mathcal{P}_I$, attribute prompts $\mathcal{P}_A$, and textual language prompts $\mathcal{P}_T$), and the latter includes a cycle alignment network. Our proposed framework can be formulated as:
\begin{equation}
\footnotesize
\begin{aligned}
&\begin{array}{cc}
\textbf{UPT}  
\end{array}
\left\{
\begin{array}{ll}
\text{Visual Prompts:}  & I \to \mathcal{P}_I \to T \\
\text{Attribute Prompts:}  & A \to \mathcal{P}_A \to T \\
\text{Language Prompts:}  & T \to \mathcal{P}_T \to T 
\end{array}
\right. 
\\
\footnotesize
&\begin{array}{cc}
\textbf{MPT}  
\end{array}
\left\{
\begin{array}{ll}
\text{Cycle Alignment:}  & \{\mathcal{P}_I, \mathcal{P}_A, \mathcal{P}_T\} \to \hat{\mathcal{P}}  \\
\text{Aligned Prompts:} & \{I, A\} \to \hat{\mathcal{P}} \to T 
\end{array}
\right.  
\end{aligned}
\end{equation}
The prompts across different modalities are used to learn the novel product domain characteristics from the limited available data in the UPT and then are used by the cycle alignment network to highlight and capture the important characteristics $\hat{\mathcal{P}}$, which is retrieved by the image and attributes to learn to generate novel product titles $T$ in the MPT.
We adopt the ViT \cite{he2016deep} from CLIP \cite{Radford2021CLIP} as the image encoder and the BERT \cite{Devlin2019BERT} from CLIP \cite{Radford2021CLIP} as the attribute/text encoder. 
For the text decoder, we adopt the Transformer-BASE \cite{Vaswani2017Transformer,fenglin2020rskip}.
In particular, CLIP and Transformer have shown great success in bridging/aligning multi-modalities \cite{nukrai2022text} and image-based natural language generation \cite{Cornia2020M2}, respectively.
During inference, we directly follow the $\{I, A\} \to \hat{\mathcal{P}} \to T$ pipeline to generate final novel product titles.

\subsection{Multimodal Prompt Learning}
When encountering a new product, the deep learning model usually suffers from significant performance degradation \cite{Alyafeai2020survey_transfer,Pan2010survey_transfer,Zhuang2021survey_transfer}, which is caused by the new domain characteristics and new writing styles of the novel product.
Therefore, to efficiently train and deploy the data-driven deep learning models on a few samples of novel products, we propose the Multimodal Prompt Learning framework, consisting of a Unimodal Prompt Training module and a Multimodal Prompt Training module.

\subsubsection{Unimodal Prompt Training}
The module introduces visual prompts, attribute prompts, and textual language prompts to learn the novel product domain characteristics and the writing styles.
We first acquire the representations of image $R_I$, attribute  $R_A$, and title $R_T$.
Then, we build three sets of trainable soft prompts \cite{Li2021Prefix,qin2021learning,Gu2022PPT,zhou2022learning}:
visual prompts $\mathcal{P}_I$, attribute prompts $\mathcal{P}_A$, and textual language prompts $\mathcal{P}_T$.
The dimensions of different prompts are all $N_\text{P} \times d$, where $N_\text{P}$ denotes the total number of soft prompts, which are used to learn and store the new characteristics of the novel product through our method, defined as follows:
\begin{equation}
\small
\hat{\mathcal{P}}_I = [\mathcal{P}_I;R_I],
\hat{\mathcal{P}}_A = [\mathcal{P}_A;R_A],
\hat{\mathcal{P}}_T = [\mathcal{P}_T;R_T]
\end{equation}
$[\cdot;\cdot]$ denotes the concatenation operation. 
Then, the prompts of images, attributes, and titles are directly inputted to the decoder as prefixes to train the model by generating (i.e., reconstructing) the titles.
Given the ground truth $T=\{w_1,w_2,\dots,w_{N}\}$, we train the model by minimizing the widely-used natural language generation loss, i.e., cross-entropy loss, defined as follows:
\begin{equation}
\small
\begin{aligned}
L^I_{\text{XE}}&=-\sum_{t=1}^{N} \log \left(p(w_{t} \mid w_{1: t-1}; \hat{\mathcal{P}}_I, I)\right) \\
L^A_{\text{XE}}&=-\sum_{t=1}^{N} \log \left(p(w_{t} \mid w_{1: t-1}; \hat{\mathcal{P}}_A, A)\right) \\
L^T_{\text{XE}}&=-\sum_{t=1}^{N} \log \left(p(w_{t} \mid w_{1: t-1}; \hat{\mathcal{P}}_T, T)\right)
\end{aligned}
\end{equation}
Finally, by combining the $L^I_{\text{XE}}$, $L^A_{\text{XE}}$, and $L^T_{\text{XE}}$, the full training objective of the Unimodal Prompt Training process is:
\begin{equation}
\footnotesize
{L}_\text{full}= \lambda_1 L^I_{\text{XE}} + \lambda_2 L^A_{\text{XE}} + \lambda_3 L^T_{\text{XE}}
\end{equation}
where $\lambda_{1,2,3} \in [0,1]$ is the hyperparameters that controls the regularization. 
We find that our approach can achieve competitive results with the state-of-the-art models with only 1\% training data when setting $\lambda_1 = \lambda_2 = \lambda_3 = 1$, thus we do not attempt to explore other settings.

Through the above operation, our Unimodal Prompt Training process can enable the model to learn the domain characteristics and the writing styles of novel products on a small amount of data. 
It is worth noting that the auto-encoding process in $L^T_{\text{XE}}$, which reconstructs the input titles, is unsupervised.
It indicates that our method 1) can be further improved by using more large-scale unlabeled texts; 2) can control the style of the generated titles by adjusting the style of input titles; and 3) can continuously learn from newly added texts of novel products to boost the performance as novel products are developed.

\begin{figure}[t]
\centering
\includegraphics[width=1\linewidth]{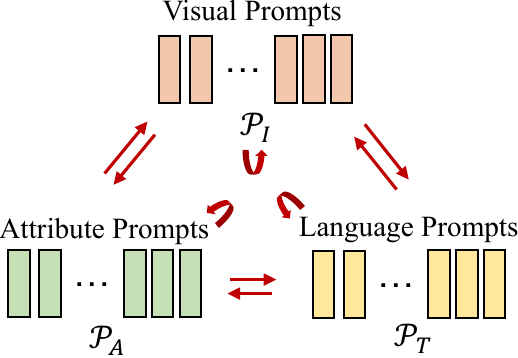}
\caption{An overview of the introduced Cycle Alignment Network. It aligns the multiple prompts across different modalities, which preserve the novel product domain characteristics, to capture the important characteristics to boost the generation of accurate and concise titles of novel products.}
\label{fig:CAN}
\end{figure}

\subsubsection{Multimodal Prompt Training}
After learning the novel domain characteristics and the new writing styles of novel products in the Unimodal Prompt Training process, we further propose the Multimodal Prompt Training process to train the framework, learning to capture the important characteristics in different prompts and describe the novel product based on the input image and attributes of the novel product.
In implementations, we first extract the representations of input image $R_I$ and input attributes $R_A$. 
Then, to boost performance, we propose to capture important characteristics and filter noisy characteristics from the visual prompts $\mathcal{P}_I$, attribute prompts $\mathcal{P}_A$, and language prompts $\mathcal{P}_T$.
Considering that important characteristics will appear in the three prompts simultaneously, we introduce the Cycle Alignment Network to perform cycle alignment of different prompts.
As shown in Figure~\ref{fig:CAN}, we take the visual prompts $\mathcal{P}_I$ as a `query' to retrieve the related novel product characteristics preserved in visual prompts $\mathcal{P}_I$, attribute prompts $\mathcal{P}_A$, and language prompts $\mathcal{P}_T$:
\begin{equation}
\footnotesize
\begin{aligned}
\mathcal{P}_{I \rightarrow I} = \boldsymbol{\alpha} \mathcal{P}_I = \sum_{k=1}^{N_\text{P}} \alpha_k p_k , \text{where} \ \boldsymbol{\alpha} = \text{softmax}(\mathcal{P}_I \mathcal{P}_I^{\top}) \\
\mathcal{P}_{I \rightarrow A} = \boldsymbol{\beta} \mathcal{P}_A = \sum_{k=1}^{N_\text{P}} \beta_k p_k , \text{where} \ \boldsymbol{\beta} = \text{softmax}(\mathcal{P}_I \mathcal{P}_A^{\top}) \\
\mathcal{P}_{I \rightarrow T} = \boldsymbol{\gamma} \mathcal{P}_T = \sum_{k=1}^{N_\text{P}} \gamma_k p_k , \text{where} \ \boldsymbol{\gamma} = \text{softmax}(\mathcal{P}_I \mathcal{P}_T^{\top}) \nonumber 
\end{aligned}
\end{equation}
Similarly, we can take the attribute prompts $\mathcal{P}_A$ and language prompts $\mathcal{P}_T$ as a `query' to retrieve the related novel product characteristics across different modalities, acquiring $\mathcal{P}_{A \rightarrow A}$, $\mathcal{P}_{A \rightarrow I}$, $\mathcal{P}_{A \rightarrow T}$, $\mathcal{P}_{T \rightarrow T}$, $\mathcal{P}_{T \rightarrow I}$, $\mathcal{P}_{T \rightarrow A}$. 
Then, we can obtain the aligned prompts $\hat{\mathcal{P}}$ by concatenating them.
Finally, given the ground truth titles $T=\{w_1,w_2,\dots,w_{N}\}$, we again adopt the cross-entropy loss to train our framework to generate the final novel product titles based on $\hat{\mathcal{P}}$:
\begin{equation}
\small
L_{\text{XE}}=-\sum_{t=1}^{N} \log \left(p(w_{t} \mid w_{1: t-1}; \hat{\mathcal{P}}, I, A)\right) .
\end{equation}
During inference, we follow the $\{I,A\} \to \hat{\mathcal{P}} \to T$ pipeline to generate titles of the test products.
In this way, our MPL framework can relax the reliance on large-scale annotated datasets and achieve competitive results with previous works with only 1\% training data.

\begin{table}[t]
\centering
\scriptsize

\begin{tabular}{@{}l|c|c@{}}
\toprule
 
Settings & Training Data
&  Testing Data   \\
\midrule [\heavyrulewidth]

Out-of-Domain & \textbf{Natural Images and Texts}   & \multirow{2}{*}{\begin{tabular}[c]{@{}c@{}} \textbf{Novel Products:} \\ PLG \\ PS \\  AF \\  IS \\ GGF  \end{tabular}} \\
\cmidrule(){1-2}
In-Domain & {\begin{tabular}[c]{@{}c@{}} \textbf{Product Images and Texts:} \\ CSJ + HK + `Electronics'+ \\ `Automotive' + SO + CPA + \\ TG + THI + OP + ACS  
\end{tabular}}   \\
\bottomrule
\end{tabular}
    \caption{Training and Testing data used for different experimental settings. We conduct the experiments on the Amazon Product Dataset \cite{Ni2019Amazon}, which consists of 15 categories of products (sorted by quantity): `Clothing Shoes and Jewelry' (CSJ), `Home and Kitchen' (HK), `Electronics', `Automotive', `Sports and Outdoors' (SO), `Cell Phones and Accessories' (CPA), `Toys and Games' (TG), `Tools and Home Improvement' (THI), `Office Products' (OP), `Arts Crafts and Sewing' (ACS), `Patio Lawn and Garden' (PLG), `Pet Supplies' (PS), `Amazon Fashion' (AF), `Industrial and Scientific' (IS), `Grocery and Gourmet Food' (GGF).}
    \label{tab:settings}
\end{table}

\section{Experiments}
In this section, we first describe a large-scale dataset, the widely-used metrics, and the settings used for evaluation.
Then, we present the results of in-domain and out-of-domain experiments.

\subsection{Datasets, Metrics, and Settings}
\myparagraph{Datasets}
We evaluate our proposed framework on a publicly available dataset, i.e., Amazon Product Dataset \cite{Ni2019Amazon}, which consists of around 15M products.
For data preparation, we first exclude entries without images/attributes/titles, which results in around 5.2M products across 15 categories.
The detailed statistics are summarized in the supplementary material. We randomly partition the dataset into 70\%-20\%-10\% train-validation-test partitions according to products.
Therefore, there is no overlap of products between train, validation, and test sets.

\begin{table*}[t]
\centering
\scriptsize
\setlength{\tabcolsep}{3.2pt}

\begin{tabular}{@{}c l c c c c c c c c c c c c c c c c @{}}
\toprule
\multirow{2}{*}[-3pt]{Settings} & \multirow{2}{*}[-3pt]{Methods} & \multirow{2}{*}[-3pt]{\begin{tabular}[c]{@{}c@{}} Ratio of \\ Data \end{tabular}} & \multicolumn{3}{c}{\textbf{PLG}} & \multicolumn{3}{c}{\textbf{PS}} & \multicolumn{3}{c}{\textbf{AF}} & \multicolumn{3}{c}{\textbf{IS}} & \multicolumn{3}{c}{\textbf{GGF}}  \\ 
\cmidrule(lr){4-6} \cmidrule(lr){7-9} \cmidrule(lr){10-12} \cmidrule(lr){13-15}  \cmidrule(lr){16-18}  
& & & B-4 & R-L & C & B-4 & R-L & C& B-4 & R-L & C& B-4 & R-L & C& B-4 & R-L & C \\
\midrule

\multirow{5}{*}{\rotatebox{90}{\begin{tabular}[c]{@{}c@{}}Supervised \\ Learning\end{tabular}}} & X-Transformer \cite{Pan2020XLinear} & 100\% & 9.8 & 17.6 & 20.5 & 9.5 & 16.0 & 22.9 & 8.1 & 13.8 & 17.8  & 6.5 & 11.7 & 13.4 & 5.4 & 8.1 & 11.5 \\
& M2-Transformer \cite{Cornia2020M2} & 100\% &  10.3 & 18.4 & 22.0 & 9.7 & 16.6 & 23.3 & 8.4 & 14.5 & 19.5 & 6.6 & 11.4 & 13.0 & 5.2 & 7.8 & 10.4   \\
&  KOBE \cite{chen2019towards} & 100\% & \bf\color{blue} 12.1 & 20.4 & 25.0  & \bf\color{blue} 11.4 & 19.7 & 26.1&  10.0 & 17.9 & 22.9 & 7.1 & 13.0 & 15.3 & 6.0 & 10.6 & 13.3  \\
& LVP-M$^3$ \cite{Guo2022LVP} & 100\% &  11.3 & 19.7  & 23.0 & 10.7 & 20.5 & 26.8 &10.2 & 18.4 & 23.6 & 7.6 & 14.1 & 16.9 & 6.5 & 11.2 & 13.4  \\ 
& CLIP-Captioning \cite{Radford2021CLIP}  & 100\% &  11.9 & \bf\color{blue} 20.9 & 24.7 &  \bf\color{blue} 11.4  &  20.3 & \bf\color{blue} 27.3 & 10.6 & 19.3 & 24.4  & 8.0 & 14.7 & 17.8 & 6.2 & 11.8 & 15.6  \\ 

\midrule

\multirow{6}{*}{\rotatebox{90}{{\begin{tabular}[c]{@{}c@{}}Few-shot \\ Learning\end{tabular}}}} 
& VL-BART \cite{Cho2021VLBART}   & \bf 1\%  & 5.9 & 11.0 & 12.2 & 6.1 & 11.0 & 12.9 & 5.7 & 9.3 & 12.3 & 5.6 & 8.7 & 9.8 & 4.7 & 7.5 & 7.8   \\
&  VL-ADAPTER \cite{Sung2022ADAPTER}   & \bf 1\% & 6.7 & 12.6 & 13.5 & 5.7 & 10.0 & 13.9  & 6.5 & 10.4 & 13.0 &  5.2 & 9.6 & 10.6 & 4.6 &7.8 & 8.7 \\
& CLIP-Captioning \cite{Radford2021CLIP}   & \bf 1\% & 7.1  & 13.2 & 15.4 & 6.2 & 10.3 & 13.4 & 6.9 & 12.0 & 13.6 & 5.6 & 9.1 & 10.9 & 5.0 & 8.2 & 8.8   \\
\cmidrule(lr){2-18}
& \multirow{2}{*}{MPL (Ours)}  & \bf 1\%  & 11.5& 20.4 & \bf\color{blue} 25.3 & 10.9 & \bf\color{blue} 20.8 & 26.1 & \bf\color{blue} 11.0 & \bf\color{blue} 20.5 & \bf\color{blue}27.7 &  \bf\color{blue} 8.9 & \bf\color{blue} 16.4 & \bf\color{blue} 19.0 & \bf\color{blue} 7.3 & \bf\color{blue} 14.7 & \bf\color{blue} 16.8    \\ 
& & 100\% & \bf\color{red} 13.5 & \bf\color{red} 22.7 & \bf\color{red} 30.6 & \bf\color{red} 12.8 &\bf\color{red}  22.0 & \bf\color{red} 29.7 & \bf\color{red} 14.1 & \bf\color{red} 24.5 & \bf\color{red} 33.3 & \bf\color{red} 10.6 & \bf\color{red} 20.1 & \bf \color{red} 23.8 & \bf \color{red} 10.1 & \bf \color{red}  18.4 & \bf \color{red} 21.9  
\\ 
\bottomrule
\end{tabular}
\caption{Results of out-of-domain experiments on five novel products (see Table~\ref{tab:settings}). B-4, R-L, and C are short for BLEU-4, ROUGE-L, and CIDEr, respectively. Higher is better in all columns. The Red- and the Blue- coloured numbers denote the best and the second-best results across all methods, respectively. } 
\label{tab:out-of-domain}
\end{table*}

\myparagraph{Metrics}
Following common practice in multimodal language generation tasks \cite{hossain2019comprehensive,specia2016shared}, we adopt the widely-used generation metrics, i.e., BLEU-4 \cite{papineni2002bleu}, ROUGE-L \cite{lin2004rouge}, and CIDEr \cite{vedantam2015cider}, which measure the match between the generated and ground truth sentences.

\myparagraph{Implementations}
We follow the state-of-the-art method CLIP \cite{Radford2021CLIP}, which has shown great success on various multimodal tasks. 
Therefore, we adopt CLIP as our base model.
In particular, the ViT \cite{Dosovitskiy2021ViT} is used as the image encoder, the BERT \cite{Devlin2019BERT} is used as the attribute/text encoder, and the Transformer-BASE \cite{Vaswani2017Transformer} is used as the text decoder.
The model size $d$ is set to 512.
Based on the average performance on the validation set, the number of prompts $N_\text{P}$ is set to 16.
For optimization, we adopt the AdamW optimizer \cite{loshchilov2019decoupled} with a batch size of 128 and a learning rate of 1e-4. We perform early stopping based on CIDEr. We apply a beam search of size 3 for inference.
Our framework is trained on 4 V100 GPUs using mixed-precision training \cite{Micikevicius2018Mixed}.

\myparagraph{Settings}
As shown in Table~\ref{tab:settings}, we perform the out-of-domain and in-domain experiments.
\begin{itemize}
    \item \textit{Out-of-Domain Experiments} are conducted by directly transferring the CLIP pre-trained on natural images and texts datasets, such as MSCOCO \cite{chen2015microsoft}, WIT \cite{Deng2009ImageNet}, and Conceptual Captions \cite{radu2018conceptual}, to the novel products.

    \item \textit{In-Domain Experiments} are conducted by pre-training the models on the top ten products in terms of quantity and then testing on the remaining five novel products. Therefore, there is no overlap of products between training and testing sets.
\end{itemize}
To improve the evaluation significantly, we further re-implement five state-of-the-art fully-supervised multimodal language generation methods, i.e., KOBE \cite{chen2019towards}, CLIP-Captioning \cite{Radford2021CLIP}, M2-Transformer \cite{Cornia2020M2}, X-Transformer \cite{Pan2020XLinear}, and LVP-M$^3$\cite{Guo2022LVP}, in which the KOBE is specifically designed for E-commerce, and two previous few-shot learning methods, i.e., VL-BART \cite{Cho2021VLBART} and VL-ADAPTER \cite{Sung2022ADAPTER}, in our experiments.

\subsection{Out-of-Domain Results}

The results are reported in Table~\ref{tab:out-of-domain}, which shows the superior performance of our approach.
As we can see, our framework outperforms previous few-shot learning methods by an average of 3.76\% BLEU-4, 7.9\% ROUGE-L, and 10.46\% CIDEr scores.
Therefore, our MPL framework not only significantly outperforms previous few-shot learning methods, but also achieves competitive results with existing state-of-the-art fully-supervised methods trained on 100\% training data with 1\% training data.
It enables our framework to provide a solid bias for novel product title generation, helping sellers save time in deploying new products.
As a result, with full training data, our method achieves the best results across different novel products.
The performances prove the validity of our method in learning the domain characteristics and the writing styles of novel products, thus relaxing the dependency on the training data to generate accurate titles for novel products with lesser annotated data.

\begin{table*}[t]
\centering
\scriptsize
\setlength{\tabcolsep}{3.2pt}
 
\begin{tabular}{@{}c l c c c c c c c c c c c c c c c c @{}}
\toprule
\multirow{2}{*}[-3pt]{Settings} & \multirow{2}{*}[-3pt]{Methods} & \multirow{2}{*}[-3pt]{\begin{tabular}[c]{@{}c@{}} Ratio of \\ Data \end{tabular}} & \multicolumn{3}{c}{\textbf{PLG}} & \multicolumn{3}{c}{\textbf{PS}} & \multicolumn{3}{c}{\textbf{AF}} & \multicolumn{3}{c}{\textbf{IS}} & \multicolumn{3}{c}{\textbf{GGF}}  \\ 
\cmidrule(lr){4-6} \cmidrule(lr){7-9} \cmidrule(lr){10-12} \cmidrule(lr){13-15}  \cmidrule(lr){16-18}  
& & & B-4 & R-L & C & B-4 & R-L & C& B-4 & R-L & C& B-4 & R-L & C& B-4 & R-L & C \\
\midrule

\multirow{5}{*}{\rotatebox{90}{\begin{tabular}[c]{@{}c@{}}Supervised \\ Learning\end{tabular}}} & X-Transformer \cite{Pan2020XLinear} & 100\% & 12.1 & 22.0 & 27.1 & 12.4 & 21.3 & 29.0 & 11.5 & 19.9 & 27.8 & 8.5 & 16.1 & 17.3 & 6.0 & 10.6 & 13.9 \\
& M2-Transformer \cite{Cornia2020M2} & 100\%  & 12.5 & 21.4 & 26.7 & 12.1 & 20.6 & 28.8 & 11.4 & 20.6 & 28.1 & 8.9 & 16.7 & 18.5 & 6.2 & 11.5 & 14.0  \\
&  KOBE \cite{chen2019towards} & 100\% & 13.9 & 22.8 & 30.6 & 15.8 & 25.9 & 35.0 & 13.2 & 21.5 & 31.0 & 9.8 & 17.3 & 20.1 & 7.5 & 14.7  &  16.2 \\
& LVP-M$^3$ \cite{Guo2022LVP} & 100\% & 13.4 & 21.9 & 30.1 & 14.2 & 24.8 & 33.7 & 14.0 & 23.1 & 31.8 & 10.1 & 17.9 & 20.6 & 8.0 & 16.3 & 17.7 \\ 
& CLIP-Captioning \cite{Radford2021CLIP}  & 100\% & \bf\color{blue} 14.2 & \bf\color{blue}  23.5 & \bf\color{blue}  31.7 & \bf\color{red}15.0 & \bf\color{red} 25.2 & \bf\color{blue}  34.6 & \bf\color{blue}  13.9 & \bf\color{blue}  23.6 & 32.3 & \bf\color{blue}10.4 & \bf\color{blue} 18.7 & \bf\color{blue}21.8 & 8.5 & 16.6 & 18.7  \\

\midrule 

\multirow{6}{*}{\rotatebox{90}{{\begin{tabular}[c]{@{}c@{}}Few-shot \\ Learning\end{tabular}}}} 
& VL-BART \cite{Cho2021VLBART}   & \bf 1\% & 6.5 & 12.3 & 14.4 & 6.8 & 12.5 & 14.2 & 6.6 & 10.9 & 13.3 & 6.5 & 11.0 & 12.5 & 5.1 & 9.8 & 12.4  \\
&  VL-ADAPTER \cite{Sung2022ADAPTER}   & \bf 1\% & 7.4 & 14.0 & 15.4 & 6.7 & 12.2 & 14.7 & 6.9 & 11.5  & 14.1 & 6.6 & 11.0 & 12.9 & 5.8 & 10.3 & 12.9 \\
& CLIP-Captioning \cite{Radford2021CLIP}   & \bf 1\% & 7.5 & 13.7 & 16.0 & 7.1 & 12.9 & 15.1 & 7.5 & 12.9 & 14.5 & 7.0 & 11.2 & 13.3 & 6.2  & 10.7 & 13.0   \\
\cmidrule(lr){2-18}
& \multirow{2}{*}{MPL (Ours)}  & \bf 1\%  & 12.6 & 22.4 & 27.0 & 12.9 & 23.3 &  30.1 & 13.4 & 23.5 & \bf\color{blue} 
 32.5 & 9.7 & 17.4 & 20.5 & \bf\color{blue} 8.8 & \bf\color{blue} 17.1 & \bf\color{blue} 19.2     \\ 
& & 100\% & \bf\color{red}  14.9 & \bf\color{red} 24.0 & \bf\color{red} 32.5 & \bf\color{blue}  14.6 & \bf\color{blue}  24.9  & \bf\color{red}  35.0 & \bf\color{red} 15.3 & \bf\color{red} 24.7 & \bf\color{red} 34.2 & \bf\color{red} 13.5 & \bf\color{red} 23.8 & \bf\color{red} 27.4 &  \bf\color{red}11.0  &  \bf\color{red} 19.5  & \bf\color{red} 23.6
\\ 
\bottomrule
\end{tabular}
\caption{In-domain experiments of our approach.  With only 1\% downstream labelled data for training, MPL can achieve competitive results with previous state-of-the-art fully-supervised methods trained on 100\% training data.}
\label{tab:in-domain}
\end{table*}

\subsection{In-Domain Results}

Table~\ref{tab:in-domain} shows that under the in-domain setting, with only 1\% training data, our MPL framework can surpass several state-of-the-art fully-supervised methods, e.g., X-Transformer \cite{Pan2020XLinear} and M2-Transformer \cite{Cornia2020M2}, and significantly outperforms previous few-shot methods across all products on all metrics.
Meanwhile, with 100\% training data as in previous works, our approach achieves average 1.46\%, 1.86\%, and 2.72\% absolute margins to current best results produced by CLIP \cite{Radford2021CLIP} in terms of BLEU-4, ROUGE-L, and CIDEr, respectively.
The best results validate the effectiveness of our approach in producing higher-quality product titles, under both the few-shot and supervised experimental settings, verifying its generalization capabilities.


\section{Analysis}
In this section, we conduct several analyses under the out-of-domain setting to better understand our proposed approach,

\begin{table*}[t]
\centering
\footnotesize
\begin{tabular}{@{}c l c c c c c c c c c c@{}}
\toprule

& \multicolumn{1}{c}{\multirow{2}{*}[-3pt]{Settings}} &  \multicolumn{3}{c}{UPT} & MPT & \multicolumn{3}{c}{\textbf{Few-shot (1\%)}}  & \multicolumn{3}{c}{\textbf{Supervised (100\%)}}     \\ 
\cmidrule(lr){3-5}  \cmidrule(lr){6-6}  \cmidrule(l){7-9} \cmidrule(l){10-12} 
 & & $\mathcal{P}_I$ & $\mathcal{P}_A$ & $\mathcal{P}_T$ &   Cycle Alignment & B-4  & R-L & C & B-4  & R-L & C  \\
\midrule [\heavyrulewidth]

& Base & & & &  &  5.0 & 8.2 & 8.8 & 6.2 & 11.8 & 15.6  \\
\cmidrule(l){2-12}

& (a) & $\surd$ & & & &  5.4 & 8.9 & 10.5 & 8.0 & 14.8 & 18.0   \\

&(b) &  & $\surd$ & & & 5.6 & 9.4 & 10.7 & 6.8 & 12.9  & 16.3 \\

&(c) &   & & $\surd$  & &  6.0 & 10.7 & 12.6 & 7.3 & 13.5 & 16.7  \\
\cmidrule(l){2-12} 

&(d) &  $\surd$   &  $\surd$ & $\surd$  & &  6.4 & 12.9 & 13.5 & 8.4 & 15.6 & 19.2 \\

&MPL  & $\surd$ & $\surd$  & $\surd$  & $\surd$ & \bf 7.3 & \bf 14.7 & \bf 16.8 & \bf  10.1 & \bf 18.4& \bf  21.9  \\

\bottomrule
\end{tabular}
\caption{Ablation study under the out-of-domain setting.
The Base model denotes the model directly trained on the target novel product data.
Our proposed MPL framework introduces two major components: Unimodal Prompt Training (UPT) and Multimodal Prompt Training (MPT), 
where the former includes three unimodal prompts (i.e., visual prompts, attribute prompts, and language prompts) and the latter includes a Cycle Alignment Network.}
\label{tab:ablation}

\end{table*}

\begin{figure*}[t]
\centering
\centerline{\includegraphics[width=0.996\linewidth]{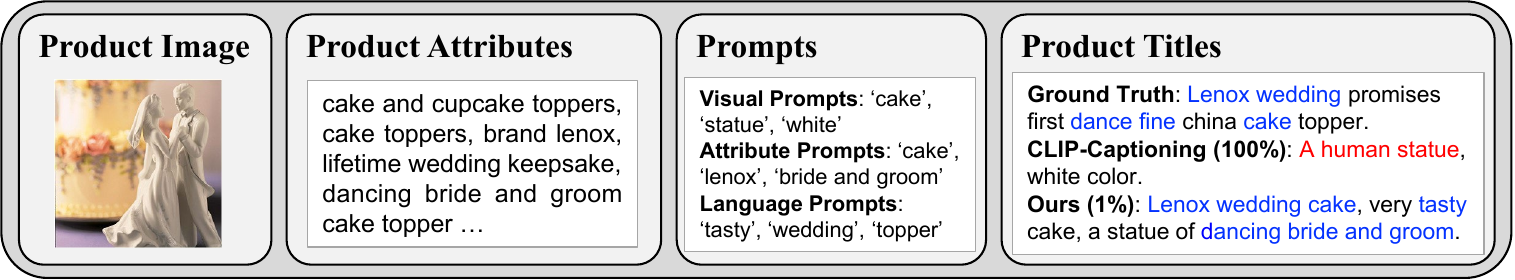}}
\caption{Novel product titles generated by the state-of-the-art fully-supervised method CLIP-Captioning \cite{Radford2021CLIP} and our approach.
Blue-colored text denotes alignment between the ground truth text and the generated text.
Red-colored text denotes unfavorable results.
We also visualize the preserved characteristics of novel products in our different prompts (top-3 attended prompts during inference).}
\label{fig:example}
\end{figure*}

\subsection{Ablation Study}
We perform the ablation study of our MPL framework to show how our approach achieves competitive results with previous works with only 1\% training data.
The results in Table~\ref{tab:ablation} show that our unimodal prompt training and multimodal prompt training of the framework all contribute to improved performances. 
It proves our arguments and the effectiveness of each proposed component.
In detail, by comparing (a-c) and Base, we can observe that the language prompts lead to the best improvements in the few-shot learning setting.
It may be explained by the fact that the language prompts $\mathcal{P}_T$ are used to reconstruct the original same input sentence, it is straightforward for the model to be trained through auto-encoding to learn the necessary domain characteristics and the writing styles using a small amount of data in the few-shot setting.
Meanwhile, the visual prompts $\mathcal{P}_I$ lead to the best improvements in the supervised learning setting.
It means that when the training data is sufficient, it is important to further capture accurate and rich visual information from the product's image to generate a desirable and concise title.
We observe an overall improvement in setting (d) by combining the three unimodal prompts, which can improve performance from different perspectives.
Table~\ref{tab:ablation} (d) and MPL show that the MPT, which includes a cycle alignment network, can bring improvements on all metrics.
It proves the effectiveness of highlighting and capturing important characteristics by aligning prompts across multiple modalities to improve performances under both few-shot and supervised settings.

\vspace{-1pt}
\subsection{Qualitative Analysis}
Figure~\ref{fig:example} gives an example to better understand our method. 
As shown in the Blue-colored text, our method is significantly better aligned with ground truth than CLIP.
For example, our framework correctly describes the key characteristics, e.g., the brand name ``\textit{Lenox}'' and the category ``\textit{wedding cake}'', and advantages, e.g., ``\textit{tasty cake}''.
However, the CLIP generates several wrong words (Red-colored text) and can not well describe the products.
More importantly, the visualization of the prompts shows that our approach can accurately learn the novel product domain characteristics to boost the generation of novel product titles.
For example, the visual prompts can accurately capture the ``\textit{cake}'', especially the attribute prompts can correctly capture the brand name ``\textit{Lenox}'' and characteristics ``\textit{bride and groom}'', and the language prompts can capture the ``\textit{tasty}'' and ``\textit{wedding}'' according to the ``\textit{cake}'' and ``\textit{bride and groom}'', respectively.

Overall, it qualitatively proves that our approach can capture important domain characteristics of novel products by multimodal prompt learning. It results in achieving competitive results with the previous supervised method CLIP with only 1\% labelled data for training, which qualitatively verifies the effectiveness of our approach in novel title generation with extremely limited labels.

\section{Conclusion}
In this paper, we present the Multimodal Prompt Learning  (MPL) framework to accurately and efficiently generate titles of novel products with limited training data.
Our MPL introduces various prompts across different modalities to sufficiently learn novel domain characteristics and writing styles, which are aligned and exploited to generate desirable novel product titles.
The out-of-domain and in-domain experiments on a large-scale dataset across five novel product categories show that, with only 1\% downstream labelled data for training, our approach achieves competitive results with fully-supervised methods.
Moreover, with the full training data used in previous works, our method significantly sets the state-of-the-art performance, which proves the effectiveness of our approach and shows its potential to deploy novel products online in time to boost product sales.

\section*{Limitations}

This paper introduces the problem of few-shot novel product title generation to efficiently and accurately generate informative and appealing titles for novel products with limited labeled data. 
However, the training of our proposed model relies on the paired image-attribute-title data, which may not be easily obtained simultaneously in the real world.
Therefore, our model may not work well when high-quality image data or textual profile is missing.
The limitations could be alleviated using techniques such as knowledge distillation or self-training. 
Besides, the writing styles of the generated titles are highly correlated with the training data. Hence, it requires specific and appropriate treatment by experienced practitioners, when deploying new products online.

\section*{Ethics Statement}
We conduct the experiments on the public dataset, which is exclusively about E-commerce and does not contain any information that names or uniquely identifies individual people or offensive content. Therefore, we ensure that our paper conforms to the ethics review guidelines.

\section*{Acknowledgements}
This paper was partially supported by NSFC (No: 62176008) and Shenzhen Science \& Technology Research Program ({\small No: GXWD20201231165807007-20200814115301001}).
\nocite{youclass,you2022bootstrapping,you2022mine,you2023rethinking,you2022momentum,you2022simcvd}

\bibliography{acl2023}
\bibliographystyle{acl_natbib}

\end{document}